# NPT-Loss: A Metric Loss with Implicit Mining for Face Recognition

Syed Safwan Khalid
University of Surrey, UK
sk0047@surrey.ac.uk

Muhammad Awais
University of Surrey, UK

Chi-Ho Chan
University of Surrey, UK

Zhen-Hua Feng
University of Surrey, UK

Ammarah Farooq
University of Surrey, UK

Ali Akbari
University of Surrey, UK

Josef Kittler
University of Surrey, UK

March 5, 2021

**Abstract**

Face recognition (FR) using deep convolutional neural networks (DCNNs) has seen remarkable success in recent years. One key ingredient of DCNN-based FR is the appropriate design of a loss function that ensures discrimination between various identities. The state-of-the-art (SOTA) solutions utilise normalised Softmax loss with additive and/or multiplicative margins. Despite being popular, these Softmax+margin based losses are not theoretically motivated and the effectiveness of a margin is justified only intuitively. In this work, we utilise an alternative framework that offers a more direct mechanism of achieving discrimination among the features of various identities. We propose a novel loss that is equivalent to a triplet loss with proxies and an implicit mechanism of hard-negative mining. We give theoretical justification that minimising the proposed loss ensures a minimum separability between all identities. The proposed loss is simple to implement and does not require heavy hyper-parameter tuning as in the SOTA solutions. We give empirical evidence that despite its simplicity, the proposed loss consistently achieves SOTA performance in various benchmarks for both high-resolution and low-resolution FR tasks.

# 1 Introduction

Automated face recognition (FR) has a wide variety of applications including surveillance, access-control, health-care, advertisement etc. Owing to its significance, it is a widely studied topic in computer vision literature. Recently, deep convolution neural network (DCNN) based solutions [32, 6, 16, 29, 30, 27, 24] have seen remarkable success in FR applications and these methods have replaced the classical FR techniques altogether. Generally, all state-of-art CNN based systems rely on the following procedure: in the **training** phase, a deep CNN is trained using a large scale datasets such as CasiaWeb [36] and/or MS-Celeb1M [10]. Some preprocessing such as face detection and alignment is carried out before training and a suitable loss function, such as triplet loss [27], normalised Softmax [29], ArcFace [6] etc., is used. Once the training is complete, the loss layer is discarded and the output of the CNN (usually a 512- or a 2048-dimensional vector) is treated as the feature vector corresponding to a given input face image. In the **testing** phase, a pair of inputs is fed to the trained network and the cosine similarity of the resulting feature vectors is evaluated. If the score is greater than a given threshold than the image pair is recognised as belonging to the same identity.

From the above procedure, we note that an accurate CNN based FR system should satisfy the following conditions: **(C1)** all feature vectors belonging to the same identity should have a large cosine similarity, i.e., all features belonging to the same person must be clustered close together, in terms of angular distance[1], in the $n$-dimensional feature space. **(C2)** Feature vectors that belong to different identities should have a sufficient amount of angular separation in the feature space, to ensure discrimination between the various identities.

We can identify two different approaches in the FR literature that try to satisfy the above mentioned conditions: the first approach is to tackle the problem directly using metric losses, such as contrastive [2] and triplet loss [27]. For instance, in the case of contrastive loss, a pair of images is fed to the network and the loss function minimises the distance between the feature vectors if the pair belongs to the same identity, and maximises the distance, if the images in the pair belong to different persons. In the case of triplet loss, instead of a pair of images, a triplet of images, consisting of an anchor, a positive and a negative sample is fed to the network. The loss is designed to minimise the distance between the anchor-positive pair and maximise it for the anchor-negative pair. While these methods are direct and straightforward, they suffer from sampling issues. For instance, if a dataset has $k$ classes with each class containing $n$ samples, then we would have triplets in the order of $\mathcal{O}(n^3)$. Hence, these direct metric losses have to rely on sampling/mining strategies that try to extract the most informative pairs/triplets out of all possibilities. Firstly, it is difficult to make efficient mining strategies; secondly, even after sampling/mining the convergence is slow for these methods.

The second approach relies on proxies rather than the actual data pairs. One example of this approach is the CenterLoss [32], which evaluates the centres of features corresponding to each identity. The distance between the features and the centres is then

---
[1]The ordering of nearest neighbours remains unaffected if we sort them using cosine similarity or with angular distance

minimized. Note that if the maximum distance between a set of features and its centres is $\epsilon$, then by triangle inequality, the maximum distance between any two features, of the same set, would be less than $2\epsilon$. Hence by minimizing the distance between the features and a proxy centre vector, we are essentially minimizing the distances among feature vectors as well. Another approach in this category is the Normalised Softmax loss [29] and its variants such as SphereLoss [16], CosineLoss [30] and ArcLoss [6] etc. Once the feature vector and class weight vectors are normalised to lie on a hyper-sphere, the resulting Softmax loss tends to maximise the cosine similarity (and hence minimise the angular distances) among the features and their corresponding class vectors. Hence the class vectors play a similar role of proxy vectors as the feature-centres played in CenterLoss. Furthermore, to enforce a separation between clusters of different identities, variants of Normalised Softmax loss, such as SphereLoss, CosineLoss and ArcLoss etc., introduce the concept of a margin that indirectly enforces separation among the weight vectors of the various identities.

While the introduction of a margin in Softmax loss has been successful and hence popular in FR literature, it is still an indirect method of attaining the condition **C2** described above, and its effectiveness is explained only intuitively. In this work, we take an alternative approach to DCNN based FR, and propose a loss (entitled nearest-neighbours-proxy-triplet or **NPT loss**) that directly creates a separation between a feature and its nearest-neighbour negative proxy/class-weight vector. Focusing on the nearest-neighbour negative class has two advantages: 1) If a separation is ensured with the nearest-neighbour, it automatically guarantees a separation with all the remaining classes. 2) It is intuitively obvious that for a given input image, a large number of its hard-negative examples will be located in the nearest-neighbour identity (See Figure 1). We give theoretical evidence to support this intuition in Proposition 1 in Section 4 and also provide empirical evidence in Section 5.3. Moreover, by working with a loss that directly creates a separation between a feature and its nearest-neighbour negative class, we are able to theoretically show that minimising the proposed loss does ensure a minimum separation between all classes (See Property 2, Section 4.1). Finally, we show using a number of experiments on various SOTA benchmark datasets, that the proposed loss consistently outperforms the SOTA solutions for both high- and low-resolution FR tasks.

## 2 Related Work

There is a long list of FR solutions in computer vision literature; however, as far as FR using DCNN is concerned, Facenet [27] can be considered as the pioneering work. In [27], it is proposed to use triplet loss with batch-mining for face recognition. Some subsequent works [8, 34] focused on efficient mining/sampling strategies to improve the performance of the trained network. Instead of directly using the feature triplets, in [32], it was suggested to approximate and update the feature centres of each class in each batch and then minimise the distance of the features from their respective centres. Another line of research focused solely on Softmax based loss functions and in [25] and [29], it was suggested that constraining the features and weight vectors to lie on a hyper-sphere is an effective strategy for improving the performance of Softmax loss for FR applications. In [16], [30], [6] and some other similar works, an additive/multiplicative margin was introduced inside the normalised Softmax loss function. The introduction of a margin significantly boosts the performance of these Softmax based loss functions and the additive margin introduced in ArcFace[6] outperformed all other solutions and has become a standard baseline for FR systems. In the last couple of years, a number of variants to [6] has been suggested with marginal improvements in the accuracy. In [38], a mechanism is defined that adjusts the hyper-parameters of ArcFace in an adaptive manner. In [31], an effort is made to combine the advantages of margin and feature mining and recently in [13], curriculum-learning [1] is incorporated with ArcFace to improve the performance. Another variant of ArcFace has been suggested in [5] that utilises multiple weight vectors for each identity. While this technique does not improve performance in itself; however, it does make the training more robust to noise in the training data.

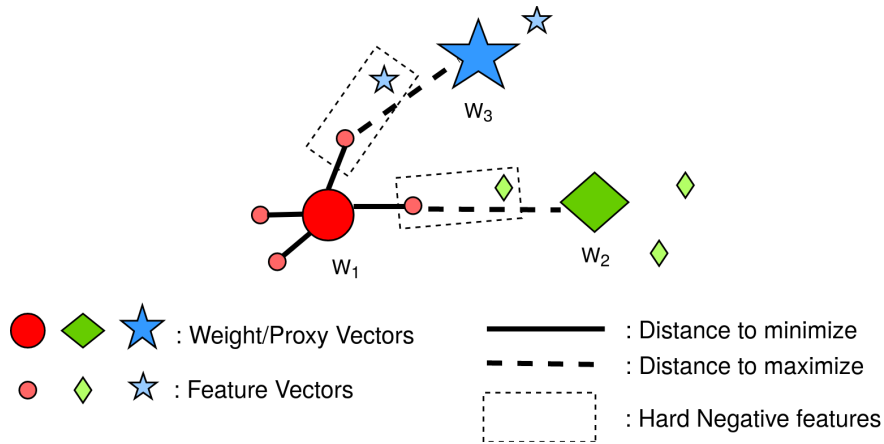

Figure 1: $W_1$, $W_2$ and $W_3$ represent class-weight vectors that can concisely represent an identity and work as proxies for the features of their respective identities. The dotted rectangles represent hard-negative pairs. The proposed loss maximises the distance of a feature with its nearest-neighbour negative proxy only, which is most likely to contain the hard negative.

It is also worth mentioning some recent developments in deep proxy based metric learning [20] that are intimately related with the modern FR solutions discussed above. In [20], proxy-based Neighbourhood Component Analysis (NCA) loss [26] and proxy-based triplet loss are discussed. Note that proxy-triplet loss also appeared independently in FR literature in [29]. The normalised Softmax is closely related with the proxy-NCA formulation and recently, [22] has shown that the normalised Softmax loss is also a smoothed version of the proxy-triplet loss. In [7] an upper bound on proxy-triplet has been suggested and optimised and in [15], an effort is made to combine pair-based and proxy-based losses.

Our work is closely related to the recent developments in both DCNN based FR and deep proxy-based metric learning. From one perspective, similar to normalised Soft-max, our proposed loss is a variant of normalised hinge loss. From another perspective, the proposed loss can be identified as a proxy-triplet loss with implicit hard-negative mining strategy. More importantly, our formulation allows us to theoretically show that minimising the proposed loss guarantees a minimum separation among all identities in the $n$-dimensional feature space. No SOTA solution has been shown to exhibit this property, to the best of our knowledge. Also empirical results show that our proposed loss consistently achieves SOTA performance which confirms its effectiveness.

## 3 Problem Formulation

Let us suppose, we have a training dataset of $N$ face images that belong to $C$ different identities. Let us denote our CNN as a nonlinear function $f(\boldsymbol{x}; \boldsymbol{w})$, where $\boldsymbol{x}$ is the input and $\boldsymbol{w}$ are the weights of the CNN. For each arbitrary input image $\boldsymbol{x}_i$ belonging to the $i$th identity, we have an output feature vector $\boldsymbol{z}_i = f(\boldsymbol{x}_i; \boldsymbol{w}) \in \mathcal{R}^n$. We define a triplet of features $(\boldsymbol{z_{i1}}, \boldsymbol{z_{i2}}, \boldsymbol{z_j})$, i.e., $\boldsymbol{z_{i1}}$ and $\boldsymbol{z_{i2}}$ belong to the same identity $i$, and $\boldsymbol{z_j}$ belongs to an identity $j \neq i$. Let $T = \{(\boldsymbol{z_{i1}}, \boldsymbol{z_{i2}}, \boldsymbol{z_j})\}$ be the set of all possible triplets from the training data. Let $d(\boldsymbol{z}_i, \boldsymbol{z}_j)$ be a distance metric defined on $\mathcal{R}^n$, then we can state the *condition of ideal ranking* as:

$$d(\boldsymbol{z}_{i1}, \boldsymbol{z}_{i2}) < d(\boldsymbol{z}_{i1}, \boldsymbol{z}_j) \qquad \forall (\boldsymbol{z_{i1}}, \boldsymbol{z_{i2}}, \boldsymbol{z_j}) \in T. \tag{1}$$

While any arbitrary distance metric might be employed in (1), in this work, we define $d(\boldsymbol{z}_i, \boldsymbol{z}_j)$ to be the squared-euclidean distance, i.e., $d(\boldsymbol{z}_i, \boldsymbol{z}_j) = (\boldsymbol{z}_i - \boldsymbol{z}_j)^T (\boldsymbol{z}_i - \boldsymbol{z}_j)$. Note that the squared-euclidean distance is not a proper metric, in the sense, that it does not obey the triangle inequality; however, the ordering of the nearest-neighbours is not affected if we use either the euclidean or the squared-euclidean distance, i.e., if $d(\boldsymbol{z}_{i1}, \boldsymbol{z}_{i2}) < d(\boldsymbol{z}_{i1}, \boldsymbol{z}_j)$, then $d_e(\boldsymbol{z}_{i1}, \boldsymbol{z}_{i2}) < d_e(\boldsymbol{z}_{i1}, \boldsymbol{z}_j)$, or vice versa, where $d_e(\boldsymbol{z}_i, \boldsymbol{z}_j) = \sqrt{d(\boldsymbol{z}_i, \boldsymbol{z}_j)}$ is the euclidean distance between $\boldsymbol{z}_i$ and $\boldsymbol{z}_j$. Since the squared-euclidean is easier to optimise and does not affect the ranking condition in (1), hence we opt for the squared distance, in our work.

It has been suggested in [29], that CNN based FR systems work better if the features are normalised to lie on a hyper-sphere. A theoretical justification is provided in [20], where it has been suggested that surrogate metric losses, such as triplet-loss etc, provide a tighter upper bound on the ideal ranking loss if features are normalised to lie on a hyper-sphere. Accordingly, in our work, we force all features $\boldsymbol{z}_i$ to lie on a hyper-sphere of radius $r$ and $d(\boldsymbol{z}_i, \boldsymbol{z}_j) = 2r^2 - 2\boldsymbol{z}_i^T \boldsymbol{z}_j = 2r^2(1 - \hat{\boldsymbol{z}}_i^T \hat{\boldsymbol{z}}_j)$, where $\hat{\boldsymbol{z}}_i$ and $\hat{\boldsymbol{z}}_j$ are unit vectors in the direction of $\boldsymbol{z}_i$ and $\boldsymbol{z}_j$, respectively, and $\hat{\boldsymbol{z}}_i^T \hat{\boldsymbol{z}}_j$ is the cosine similarity between $\boldsymbol{z}_i$ and $\boldsymbol{z}_j$.

A direct method to achieve the condition in (1) is to minimise the standard triplet loss [27], defined as:

$$L_{\text{triplet}} = \max\{0, d(\boldsymbol{z}_{i1}, \boldsymbol{z}_{i2}) - d(\boldsymbol{z}_{i1}, \boldsymbol{z}_j) + \Delta\}, \tag{2}$$

where $\Delta$ is a positive margin. However, the standard triplet loss has two serious shortcomings: firstly, for a training set of $N$ images, we will have possible triplets of the order $\mathcal{O}(N^3)$. This significantly increases the total iterations required in the training process. Secondly, most of the triplets will satisfy the triplet constraint, i.e., $d(\boldsymbol{z}_{i1}, \boldsymbol{z}_{i2}) - d(\boldsymbol{z}_{i1}, \boldsymbol{z}_j)$ will be less than $-\Delta$ for many of the triplets and hence the loss will be zero. These triplets will slow down convergence of the training process. To overcome these shortcomings, it is suggested in [27] to search/mine for *hard-positive* samples, i.e., $\max_k\{d(\boldsymbol{z}_i, \boldsymbol{z}_k)\}$, where $\boldsymbol{z}_i, \boldsymbol{z}_k$ belong to the same identity, and to search for *hard-negative* samples, i.e., $\min_j\{d(\boldsymbol{z}_i, \boldsymbol{z}_j)\}$, where $\boldsymbol{z}_i, \boldsymbol{z}_j$ belong to the separate identities. However, such a mining process is infeasible if applied across the entire training data. Since the CNN is trained using a batch stochastic gradient descent algorithm, an alternative is to mine for hard-positives and -negatives across a single batch in each iteration. This, however, necessitates that a sufficiently large batch-size is chosen, which is computationally prohibitive and also can reduce the generalisation of the trained network.

An alternative to the standard triplet loss is the proxy triplet loss suggested in [29, 20]. To apply the proxy triplet loss, we take a set of $C$, $n$-dimensional weight vectors $\boldsymbol{W}_1, \boldsymbol{W}_2, \cdots, \boldsymbol{W}_C$, with $||\boldsymbol{W}_i|| = r$, for all $i$. The weight vector $\boldsymbol{W}_i$ acts as a proxy for all features that belong to the $i$th identity. Consequently, instead of evaluating distances between sample pairs, we evaluate distances between samples and proxies, i.e., for each training sample $\boldsymbol{x}_i$, with corresponding feature $\boldsymbol{z}_i$, the following loss is evaluated:

$$L_{\text{proxy-triplet}} = \sum_j \max\{0, d(\boldsymbol{z}_i, \boldsymbol{W}_i) - d(\boldsymbol{z}_i, \boldsymbol{W}_j) + \Delta\}. \quad (3)$$

Note that, as suggested in [20], for each input $\boldsymbol{x}_i$, the proxy-triplet loss considers **all** negative proxies and hence covers the entire dataset. However, this also means that to evaluate the loss function, we consider those proxies that contain a number of hard-negative samples as well as those proxies that contain few or no hard-negative samples at all. Essentially, we are not using any mining strategy in the evaluation of proxy-triplet loss. We will show in our experiments, that the proxy-triplet loss does not work well for face recognition. One would expect that the performance of the proxy-triplet loss can be improved, if we could incorporate some mining strategy within the loss function. In the following, we suggest an alternative loss that has an implicit mechanism to mine hard-negatives and hence outperforms the standard proxy-triplet loss and achieves SOTA results.

## 4  Proposed Loss

The proposed loss relies on the concept of *nearest-neighbour negative proxy* defined as follows:

**Definition.** *Let $\boldsymbol{W}_1, \boldsymbol{W}_2, \cdots, \boldsymbol{W}_C$ be the weight vectors (i.e., proxies) corresponding to the $C$ identities. Let $\boldsymbol{z}_i$ be an arbitrary feature of identity $i$ with weight vector $\boldsymbol{W}_i$. We define $\boldsymbol{W}_j$ to be the nearest-neighbour negative proxy of $\boldsymbol{z}_i$ and denote it by $\boldsymbol{W}_n^{(i)}$, if $d(\boldsymbol{z}_i, \boldsymbol{W}_j) \leq d(\boldsymbol{z}_i, \boldsymbol{W}_k)$, for all $k \neq i$.*

Using the above definition, we define the proposed loss as follows:

$$L_{\text{NPT}} = max\{0, d(\boldsymbol{z}_i, \boldsymbol{W}_i) - d(\boldsymbol{z}_i, \boldsymbol{W}_n^{(i)}) + \Delta\}. \quad (4)$$

Note that instead of using all negative proxies, the proposed loss only uses the nearest-neighbour negative proxy, since the samples corresponding to the nearest-neighbour negative proxy are most likely to be hard-negatives (see Figure 1). This intuitive idea is further confirmed by the following proposition and its corollary:

**Proposition 1.** *Let $z_i$ be a given feature vector belonging to identity $i$. Let $z_j$ and $z_k$ be random and independent feature vectors drawn from identities $j$ and $k$, respectively, with $\mathbb{E}[z_j] = m_j$ and $\mathbb{E}[z_k] = m_k$. Let $\tilde{m}_j = r\frac{m_j}{||m_j||}$ and $\tilde{m}_k = r\frac{m_k}{||m_k||}$. Let $W_i$, $W_j$, $W_k$ be the weight vectors corresponding to $z_i$, $z_j$, and $z_k$, respectively. If the following assumptions are true:*

**A1**: $||m_j|| = \beta$, *for all* $j = 1, 2, \cdots, C$.

**A2**: $W_j = \tilde{m}_j$, *for all* $j = 1, 2, \cdots, C$.

*Then,* $\mathbb{E}[d(z_i, z_j)|z_i] < \mathbb{E}[d(z_i, z_k)|z_i]$, *if* $d(z_i, W_j) < d(z_i, W_k)$.

**Proof**: We note that $d(z_i, z_j) = 2r^2 - 2z_j^T z_i$ and $d(z_i, z_k) = 2r^2 - 2z_k^T z_i$. Taking expectation, we can write

$$\mathbb{E}[d(z_i, z_j)|z_i] = 2r^2 - 2\gamma W_j^T z_i, \tag{5}$$

and

$$\mathbb{E}[d(z_i, z_k)|z_i] = 2r^2 - 2\gamma W_k^T z_i, \tag{6}$$

where $\gamma = \frac{\beta}{r} > 0$. Subtracting (6) and (5), we get

$$\mathbb{E}[d(z_i, z_k)|z_i] - \mathbb{E}[d(z_i, z_j)|z_i] = 2\gamma(W_j - W_k)^T z_i$$
$$= \gamma \left[ (2r^2 - 2W_k^T z_i) - (2r^2 - 2W_j^T z_i) \right]$$
$$= \gamma \left[ d(z_i, W_k) - d(z_i, W_j) \right].$$

Consequently, if $d(z_i, W_j) < d(z_i, W_k)$, then $\mathbb{E}[d(z_i, z_k)|z_i] - \mathbb{E}[d(z_i, z_j)|z_i] > 0$ and hence $\mathbb{E}[d(z_i, z_j)|z_i] < \mathbb{E}[d(z_i, z_k)|z_i]$, which is the required proof.

**Corollary.** *Let $z_i$ be a given feature vector belonging to identity $i$. Let $z_j$ and $z_k$ be random and independent feature vectors drawn from identities $j$ and $k$, respectively, with $W_j = W_n^{(i)}$. Then $\mathbb{E}[d(z_i, z_j)|z_i] < \mathbb{E}[d(z_i, z_k)|z_i]$ for all $k = 1, 2, \cdots, C, k \neq i, j$.*

**Proof**: The result follows from Proposition 1 and the definition of nearest-neighbour negative proxy $W_n^{(i)}$.

**Remark 1.** From the above results, we note that, on average, the distance between $z_i$ and a sample from its nearest-neighbour negative proxy is less than the distance between $z_i$ and a sample from any other proxy. Hence, if we mine for hard-negative samples, on average, they will be found in the identity corresponding to the nearest-neighbour negative proxy. The proposed loss in (4) is enforcing a separation between $z_i$ and $W_n^{(i)}$, and hence, on average is enforcing a separation between $z_i$ and its corresponding hard-negatives. Consequently, the proposed loss is acting as a proxy-triplet loss similar to (3); however, with implicit hard-negative mining.

**Remark 2.** The above results rely on the assumptions **A1** and **A2**. **A1** states that $||\mathbb{E}[z_j]|| = ||m_j||$ is the same for all identities. To understand this assumption, note that $\mathbb{E}[d(z_j, m_j)] = r^2 - ||m_j||^2$. Hence **A1** is the assumption that the expected distance of a sample from its mean is invariant of its identity. This assumption makes sense, in terms of face recognition tasks, as we can think of the mean feature vector as containing the core identity features and the variations from the mean arising from the variations in pose, illuminations, blur, etc. Since all identities are equally likely to undergo these variations, hence we can expect that the average distance of the features from their respective means would be independent of the actual identities of the features. In Section 5.3, we also give some empirical evidence that supports **A1**. **A2** states that the class-weight/proxy vector $W_j$ is the same as the normalised mean $\tilde{m}_j$. In other words, the angular distance between the proxy and the mean is zero (or the cosine similarity is one). This is a strong assumption and only becomes true later in the training process(see Table 1). However, **A2** can be relaxed and the result in Prop. 1 still holds, as discussed below:

**Proposition 2.** *Let $z_i$, $W_j$, $W_k$, $\tilde{m}_j$ and $\tilde{m}_k$ be arbitrary vectors in $\mathcal{R}^n$. If $d(z_i, W_j) < d(z_i, W_k)$ and $d_e(W_j, \tilde{m}_j) + d_e(W_k, \tilde{m}_k) < \alpha$, where $\alpha = d_e(z_i, W_k) - d_e(z_i, W_j)$, then $d(z_i, \tilde{m}_j) < d(z_i, \tilde{m}_k)$.*

**Proof**: Using triangle inequality

$$d_e(z_i, \tilde{m}_j) \leq d_e(z_i, W_j) + d_e(W_j, \tilde{m}_j) \qquad (7)$$

Also, using reverse-triangle inequality

$$d_e(z_i, \tilde{m}_k) \geq |d_e(z_i, W_k) - d_e(W_k, \tilde{m}_k)| \qquad (8)$$

Noting that $d_e(z_i, W_k) = d_e(z_i, W_j) + \alpha$, (8) can be written as

$$d_e(z_i, \tilde{m}_k) \geq |d_e(z_i, W_j) + \alpha - d_e(W_k, \tilde{m}_k)| \qquad (9)$$

Now, if $d_e(W_j, \tilde{m}_j) + d_e(W_k, \tilde{m}_k) < \alpha$, then $\alpha - d_e(W_k, \tilde{m}_k) > d_e(W_j, \tilde{m}_j)$, hence (9) can be written as

$$d_e(z_i, \tilde{m}_k) \geq d_e(z_i, W_j) + d_e(W_j, \tilde{m}_j) \qquad (10)$$

From (7) and (10), we get $d_e(z_i, \tilde{m}_j) < d_e(z_i, \tilde{m}_k)$. Taking square on both sides of the inequality we get $d(z_i, \tilde{m}_j) < d(z_i, \tilde{m}_k)$, which is the required proof.

**Corollary.** *Let A2 be false in Proposition 1. The results in Proposition 1 still hold if $\max\{d_e(W_j, \tilde{m}_j) + d_e(W_k, \tilde{m}_k)\} < \frac{\alpha}{2}$, where $\alpha = d_e(z_i, W_k) - d_e(z_i, W_j)$.*

**Proof**: Follows directly from Proposition 1 and 2.

**Remark 3.** We note from the above results that we do not require the proxy to be strictly equal to the normalised mean. As long as the distance between the proxy and the normalised mean is small, the results in Proposition 1 still hold. In Section 5.3, we will give some empirical evidence that supports Proposition 1. We will also show that the distance between the proxy and the normalised mean is indeed small and gets smaller as the training proceeds.

## 4.1 Properties of the Proposed Loss

In addition to the motivations described above, in this section we discuss some properties of the proposed loss that further support the suitability of the proposed loss for face recognition tasks.

**Property 1.** If $L_{\text{proposed}} < \Delta$ for a given $z_i$, then $d(z_i, W_i) < d(z_i, W_j)$ for all $j = 1, 2, \cdots, C, j \neq i$.

**Proof**: If $L_{\text{proposed}} < \Delta$, then $d(z_i, W_i) - d(z_i, W_n^{(i)}) < 0$, hence $d(z_i, W_i) < d(z_i, W_n^{(i)}) < d(z_i, W_j)$, where the last inequality follows from definition of $W_n^{(i)}$.

**Remark 4.** Note that if the above property is true for all training samples $\{z_i\}$, then we get ideal classification. However, in face recognition, we not only wish to achieve ideal classification but also require separation between classes. The proposed loss ensures this separation, as indicated by the property below:

**Property 2.** If $\mathbb{E}[L_{\text{proposed}}] < \epsilon$, where $\epsilon$ is a small positive number and assumptions **A1** and **A2** are true, then $d(W_i, W_j) > \Delta - \epsilon$ for all $i, j = 1, 2, \cdots, C, i \neq j$.

**Proof**: If $\mathbb{E}[L_{\text{proposed}}] < \epsilon$, then $\mathbb{E}[d(z_i, W_n^{(i)})] - \mathbb{E}[d(z_i, W_i)] > \Delta - \epsilon$. Noting that $\mathbb{E}[d(z_i, W_i)] = 2r^2 - 2\gamma r^2$ and $\mathbb{E}[d(z_i, W_n^{(i)})] = 2r^2 - 2\gamma W_i^T W_n^{(i)}$, we can write

$$\begin{aligned}\mathbb{E}[d(z_i, W_n^{(i)})] - \mathbb{E}[d(z_i, W_i)] &= \gamma(2r^2 - 2W_i^T W_n^{(i)}) \\ &= \gamma d(W_i, W_n^{(i)}) > \Delta - \epsilon\end{aligned} \quad (11)$$

Note that $\gamma = \frac{\beta}{r}$, where $\beta = ||\mathbb{E}[z_i]||$. Since $\mathbb{E}[d(z_i, \mathbb{E}[z_i])] = r^2 - \beta^2 > 0$, hence $\beta < r$ and $0 < \gamma < 1$. Consequently, $d(W_i, W_n^{(i)}) > \frac{\Delta - \epsilon}{\gamma} > \Delta - \epsilon$. Also, by definition of $W_n^{(i)}$, $d(W_i, W_n^{(i)}) \leq d(W_i, W_j)$ for $j \neq i$, hence we can write $d(W_i, W_j) > \Delta - \epsilon$, which is the required proof.

**Remark 5.** Note that at the end of the training process, $\mathbb{E}[L_{\text{proposed}}] \approx 0$ and hence $d(W_i, W_j)$ is guaranteed to be greater that $\Delta$. Therefore, minimizing the proposed loss ensures that all classes are at-least separated by a margin $\Delta$. Ideally, we would like $\Delta$ to be as large as possible; however, our empirical results show that any value of $\Delta$ greater than $\frac{r^2}{2}$ increases the steady-state value of the expected loss and does not yield any performance improvements. Note that, for instance, if we put $\Delta = \frac{r^2}{2}$ in (4), we can write

$$L_{\text{NPT}} = 2r^2 \max\{0, \hat{z}_i^T \hat{W}_n^{(i)} - \hat{z}_i^T \hat{W}_i + \frac{1}{2}\}, \quad (12)$$

where $\hat{z}_i$, $\hat{W}_n^{(i)}$ and $\hat{W}_i$ are unit vectors in the direction of $z_i$, $W_n^{(i)}$ and $W_i$, respectively. The radius of the hyper-sphere acts as a multiplication factor in the above formulation and hence it can be lumped in the learning rate. In this manner, the proposed loss becomes hyper-parameter free, i.e., the only hyper-parameters that require tuning are the learning rate, weight decay and momentum etc., that are involved in all CNN based training and the proposed loss does not introduce any new hyper-parameters. This is in contrast to ArcFace that requires two extra parameters to be tuned for each training process.

# 5 Experiments

## 5.1 Datasets

For training, we work with the CASIA (i.e., small protocol) [36] and the MS1M (i.e., large protocol) [10] datasets. Both datasets are available from the insightFace repository [3] and we use the ArcFace version of MS1M, i.e., MS1Mv2. CASIA contains around 10K identities and 0.5M images; whereas, MS1Mv2 contains about 85K identities and around 5.8M images. For testing, we use the standard benchmark LFW[11], CPF-FP [28] for FR under pose variations, CALFW [40] and AgeDB[19] for FR under age variations, and large scale datasets such as MegaFace [14], IJBB [33], IJBC [18] for FR under realistic unconstrained environments [2]. We also give results on SCFace [9] dataset that contains low-resolution face images.

## 5.2 Implementation Details

We perform all experiments on a ResNet-50 architecture described in Figure 2, trained using PyTorch [21]. For CASIA, we employ a batch size of 64 and a base learning rate of $0.1$ that is divided by 10, at 30th and 45th epoch. The training is finished after 65 epochs. For MS1M, we use a batch size of $512$, and a base learning rate of $0.1$ that is divided by 10, at 10th and 18th epoch. The training is finished after 21 epochs. The momentum and weight decay are the same for both datasets., i.e., $0.9$ and $1e^{-4}$, respectively.

## 5.3 Empirical evidence regarding Propositions

In Proposition 1, in Section 4, we showed that, under mild assumptions, the expected distance of a given sample from the samples of its nearest-neighbour negative proxy is less than its distance from samples of all other proxies. Here we give empirical evidence to support this proposition and also give some evidence to support the assumptions **A1** and **A2** that are used in our work.

We take take all classes from the training data (i.e., MS1M) with samples $> 500$ and evaluate the features using the trained ResNet50 model. For each sample $z_i$, we find class-weight vectors $W_j$ and $W_k$, i.e., $d(z_i, W_j) \leq d(z_i, W_k) \leq d(z_i, W_l)$ for all $j, k, l \neq i$. Note that $W_j$ is the nearest-neighbour negative class for $z_i$ and $W_k$ is the next nearest class after $W_j$. We then approximate $\mathbb{E}[d(z_i, z_j)] = d_{n,i} \approx \frac{1}{N_j} \sum_j d(z_i, z_j)$ and $\mathbb{E}[d(z_i, z_k)] = d_{k,i} = \frac{1}{N_k} \sum_j d(z_i, z_k)$, where $N_j$ and $N_k$ are number of samples in class $j$ and $k$, respectively. We evaluate $D_n = \frac{1}{N_C} \sum_i d_{n,i}$ and $D_k = \frac{1}{N_C} \sum_i d_{k,i}$, where $N_C$ is the total number of classes in this experiment. We plot $D_n$ and $D_k$, in Figure 3, and note that, $D_n < D_k$ for all epochs in our experiment, as predicted by Proposition 1.

Assumption **A1** states that $||m_i|| = ||\mathbb{E}[z_i]||$ is the same for all $i$. We approximate $||\mathbb{E}[z_i]|| \approx ||\frac{1}{N_i} \sum_i z_i||$ and tabulate the variance in the values of the various classes in

---

[2]For an independent evaluation, we have also submitted our trained network in the IFRT [4] challenge; we are awaiting results and shall include them in the supplementary material.

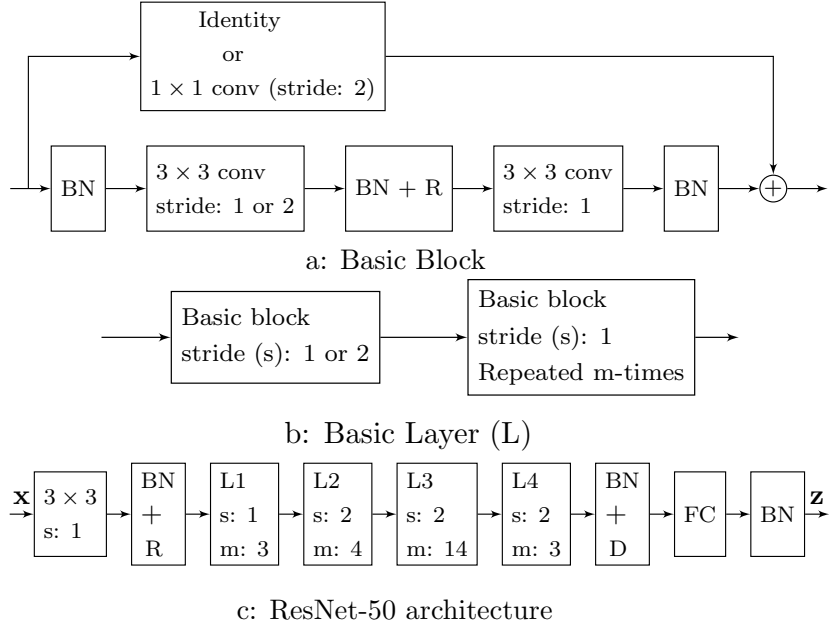

a: Basic Block

b: Basic Layer (L)

c: ResNet-50 architecture

Figure 2: (a) Basic building block of ResNet-50. The residual connection is identity if the first conv layer in the basic block has stride 1; otherwise, a $1 \times 1$ conv with stride 2 is used. (b) Basic layer: first basic block has stride 1 or 2 and then an m-times repetition of the basic block with stride 1. (c) The overall architecture: BN is batch-norm, R is relu, D is dropout and FC is the fully connected layer. Input **x** is a $112 \times 112$ image and output **z** is a $512$-dim feature vector.

Table 1. Note that ideally, the variance should be zero; however, since we are working with a finite number of samples, we would expect a small but non-zero variance in the approximated values of $||\boldsymbol{m}_i||$. We note from Table 1 that the variance is small, for all epochs, which supports **A1**. It is also interesting to note the vale of $\gamma = \frac{\beta}{r} = \frac{||\boldsymbol{m}_i||}{r}$, assumed to be same for all $i$, which we approximate as $\gamma \approx \bar{\gamma} = \frac{1}{N_C} \sum_i \frac{||\boldsymbol{m}_i||}{r}$ and tabulate it in Table 1. We note that as the training proceeds, $\bar{\gamma}$ becomes closer to 1 and since $\mathbb{E}[d(\boldsymbol{z}_j, \boldsymbol{m}_j)] = r^2 - ||\boldsymbol{m}_j||^2$, hence $\mathbb{E}[d(\boldsymbol{z}_j, \boldsymbol{m}_j)]$, i.e, the distance between features from their respective means, becomes close to zero. This shows that the proposed loss is enforcing intra-class compactness, which is a desirable result for FR tasks (See condition **C1** in Section 1).

Assumption **A2** states that $\boldsymbol{W}_i$ and $\tilde{\boldsymbol{m}}_i$ are equal, where $\boldsymbol{W}_i$ is the class-weight vector of identity $i$ and $\tilde{\boldsymbol{m}}_i = \frac{\mathbb{E}[\boldsymbol{z}_i]}{||\mathbb{E}[\boldsymbol{z}_i]||}$ is the normalised mean. Since both $\boldsymbol{W}_i$ and $\tilde{\boldsymbol{m}}_i$ lie on the same hyper-sphere, hence **A2** is equivalent to the statement that the $\text{Cos}(\boldsymbol{W}_i, \tilde{\boldsymbol{m}}_i)$ is 1. Note that we relaxed this assumption using Proposition 2 and only require $\boldsymbol{W}_i$ and $\tilde{\boldsymbol{m}}_i$ to be close to each other. The average cosine similarity between $\boldsymbol{W}_i$ and $\tilde{\boldsymbol{m}}_i$ is Tabulated in Table 1, Note that, both $\boldsymbol{W}_i$ and $\tilde{\boldsymbol{m}}_i$ are close to each other by the end of Epoch-3 and by the end of training $\text{Cos}(\boldsymbol{W}_i, \tilde{\boldsymbol{m}}_i) \approx 1$, which justifies

A2.

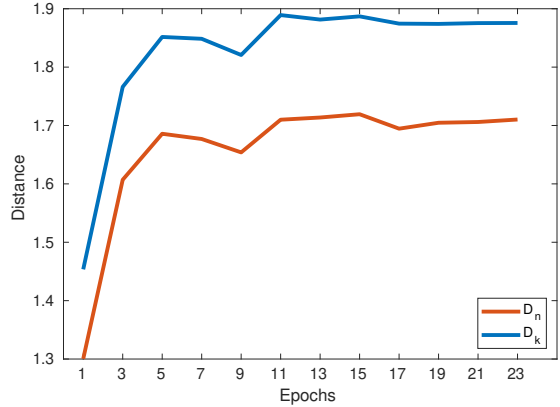

Figure 3: Comparison of average distance of a sample from features of its nearest-neighbour negative class vs., other classes.

| Epochs | $\bar{\gamma}$ | Variance in $\|\boldsymbol{m}_i\|$ | $\mathrm{Cos}(\boldsymbol{W}_i, \tilde{\boldsymbol{m}}_i)$ |
|---|---|---|---|
| 1 | 0.725 | 0.001 | 0.6114 |
| 3 | 0.729 | 0.002 | 0.9164 |
| 11 | 0.818 | 0.0013 | 0.9838 |
| 21 | 0.912 | 0.0004 | 0.9964 |

Table 1: Column 1: Approximated value of $\gamma$. Column 2: Variance in the values of approximate $\|\mathbb{E}[\boldsymbol{z}_i]\|$. Column 3: Avg. Cosine similarity between class-weights and feature mean vectors.

## 5.4 Ablation Study

We note in Remark 5 that the radius of the hyper-sphere can be lumped in the learning rate and the only hyper-parameter in the proposed loss in $\Delta$. We evaluate the performance of our model trained on CASIA for different values of $\Delta$ on a subset of MegaFace with 10K distractors. The rank-1 accuracy is plotted in Figure 4. We note that for all values of $\Delta > \frac{r^2}{2}$, the performance remains almost the same (see also Remark 5). Hence in all our experiments, we set $\Delta = \frac{r^2}{2}$ and perform no further parameter tuning. The lack of dependence of the NPT-loss from hyper-parameter tuning is a desirable advantage over SOTA techniques, such as ArcFace, which require hyper-parameter tuning for each new scenario.

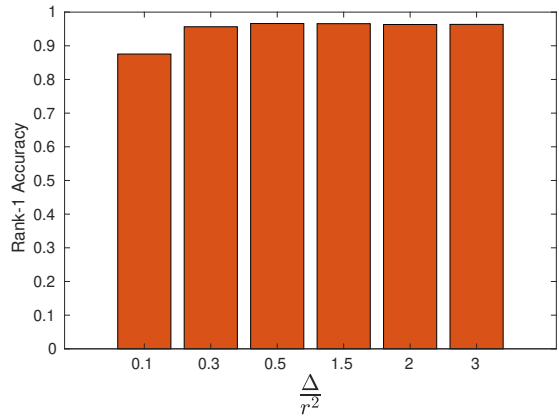

Figure 4: Comparison of rank-1 accuracy evaluated on MegaFace (10K distractors) of ResNet-50 model trained on CASIA, for different values of the hyper-parameter $\Delta$.

## 5.5 Comparison with SOTA

### 5.5.1 Small Protocol

SOTA FR solutions rarely report performance on small protocol; therefore, we trained our ResNet-50 model on CASIA using Normalised Softmax[29], Proxy-triplet[20, 29], ArcFace [6], recently proposed Curricular-Face [13] and with the proposed loss. The performance of LFW, AgeDB, CFP-FP and CALFW is tabulated in Table 2. We note that in all instances, the proposed loss outperforms the SOTA solutions. In Table 3, we list the accuracy of these solutions on the MegaFace dataset with 1M distractors. We note that the performance of the naive proxy-triplet was reasonable for the small scale datasets in Table 2; however, when the number of distractors is large, the performance has degraded significantly. On the other hand, the proposed loss has a much superior performance and outperforms the SOTA solutions. In Figure 5, we plot the ROC curve of proposed solution, evaluated on MegaFace, and compare it against ArcFace and Curricular-Face. We note that the proposed scheme is consistently better than other losses for all values of false acceptance rate (FAR). Only at very low FAR, i.e., around $1e^{-8}$ does Curricular-Face shows slight improvements over the proposed loss.

| Method | LFW | AgeDB | CFP-FP | CALFW |
|---|---|---|---|---|
| Norm-Softmax[29] | 97.55 | 87.14 | 87.15 | 88.46 |
| Proxy-Triplet[29, 20] | 97.48 | 84.15 | 90.9 | 85.31 |
| ArcFace[6] | 99.3 | 94.23 | 95.3 | 93.34 |
| Curricular-Face[13] | 99.36 | 94.18 | 95.61 | 93.34 |
| Proposed (NPT loss) | **99.49** | **94.88** | **96.44** | **93.85** |

Table 2: Performance comparison on LFW, AgeDB, CFP-FP and CALFW for ResNet-50 trained on CASIA.

| Method | Id | Veri |
|---|---|---|
| Normalised-Softmax[29] | 59.91 | 63.04 |
| Proxy-Triplet[29, 20] | 35.85 | 46.42 |
| ArcFace[6] | 89.01 | 91.83 |
| Curricular-Face[13] | 91.52 | 93.51 |
| Proposed (NPT loss) | **92.52** | **94.47** |

Table 3: Performance comparison on MegaFace for ResNet-50 trained on CASIA. Id refers to rank-1 accuracy and Veri. is face verification performance at $1e^{-6}$ FAR.

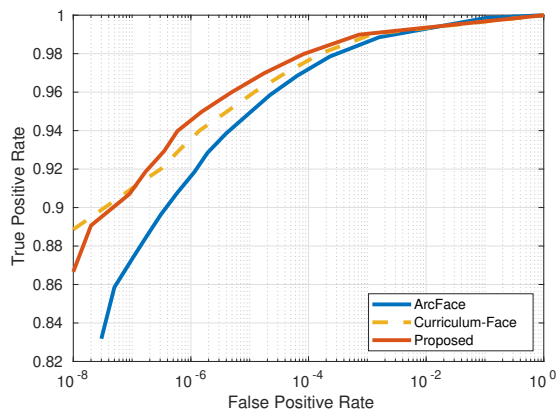

Figure 5: ROC curves evaluated on MegaFace for models trained on CASIA.

### 5.5.2 Large Protocol

In this section, we compare the performance of the proposed loss trained on MS1Mv2. We retrain ArcFace and Curricular-Face using our experimental settings; whereas, performances of other SOTA schemes have been taken from literature. In Table 4, we list the rank-1 and verification accuracy of a number of SOTA schemes evaluated on MegaFace. The verification has been performed for an FAR of $1e^{-6}$. We note that the proposed loss outperforms all SOTA solutions in both rank-1 as well as verification accuracy. In Figure 6, we plot the ROC curve of proposed solution, evaluated on MegaFace, and compare it against ArcFace and Curricular-Face. We note that all three solutions have comparable performances and the proposed scheme shows a slight improvement over other schemes at low FARs. In Table 5, we compare the performance of SOTA schemes on IJB-B and IJB-C datasets, where the results are evaluated for an FAR of $1e-4$. Similar to [13], we take the average of the image features as the corresponding template representation and apply no further strategy for set-based verification. The results show that the proposed loss achieves SOTA performance on both IJB-B and IJB-C datasets as well.

| Method | Id | Veri |
|---|---|---|
| AdaCos[38] | 97.41 | - |
| PS2Grad[39] | 97.25 | - |
| MV-Arc-SoftMax[31] | 97.14 | 97.57 |
| ArcFace[6] | 97.58 | 98.15 |
| Curricular-Face[13] | 96.98 | 98.41 |
| Proposed (NPT loss) | **97.67** | **98.54** |

Table 4: Performance comparison on MegaFace. Id refers to rank-1 accuracy and Veri. is face verification performance at $1e^{-6}$ FAR.

| Method | IJB-B | IJB-C |
|---|---|---|
| ResNet50+DCN[35] | 84.1 | 88.0 |
| CosFace[30] | - | 91.82 |
| Crystal Loss[23] | - | 92.29 |
| AdaCos[38] | - | 92.4 |
| PS2Grad[39] | - | 92.3 |
| ArcFace[6] | 93.15 | 94.79 |
| Curricular-Face[13] | 92.40 | 94.33 |
| Proposed(NPT loss) | **93.39** | **94.97** |

Table 5: Performance comparison on IJB-B and IJB-C datasets. The performance is evaluated @ FAR=$1e^{-4}$

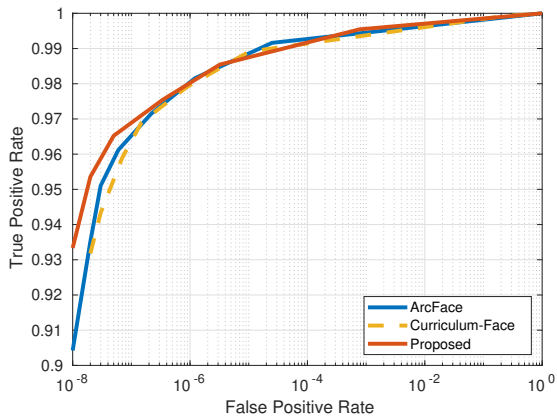

Figure 6: ROC curves evaluated on MegaFace for models trained on MS1Mv2.

### 5.5.3 Evaluation on SCFace

SOTA FR solutions work well on high resolution (HR) faces; however, the performance of these algorithms degrades significantly for low-resolution (LR) images. Since images from surveillance cameras are usually LR, it is important to evaluate the performance of an FR solution on LR images. One straight forward method to make FR solutions resolution-invariant is to train the network on HR and downsampled LR images. In this experiment, we train our ResNet-50 model using standard CASIA as well as downsampled CASIA images corresponding to face widths equal to 30, 20 and 10 pixels. The performance is tested on SCFace dataset that contains surveillance camera images taken from three different distances (i.e.,1.0m, 2.6m and 4.2m). For each identity, there is an HR mugshot image available and the standard protocol is to use mugshot HR images as gallery and the surveillance camera images as probes. The results are tabulated in Table 6. Note that the proposed solution has comparable performance with SOTA solutions for all distances and overall achieves the best average accuracy.

| Distance | d1 | d2 | d3 | avg. |
|---|---|---|---|---|
| ArcFace (Resnet50) [6, 37] | 48.0 | 92.0 | **99.3** | 79.8 |
| ArcFace-FT (Resnet50) [37] | 67.3 | 93.5 | 98.0 | 86.3 |
| DCR-FT [17] | 73.3 | 93.5 | 98.0 | 88.3 |
| FAN-FT [37] | 77.5 | 95.0 | 98.3 | 90.3 |
| Huang et.al[12] | **86.8** | 98.3 | 98.3 | 94.46 |
| Proposed (NPT loss) | 85.69 | **99.08** | 99.08 | **96.61** |

Table 6: Comparison of the Proposed scheme with state-of-the-art algorithms on SCFace.

## 6 Conclusion

In this work, we have proposed a novel loss function for face recognition that directly creates a separation between a feature vector and its nearest-neighbour negative class weight vector. We have shown that the proposed loss is equivalent to a triplet loss with proxies and an implicit mechanism of hard-negative mining. We have given theoretical evidence that minimising the proposed loss guarantees a separation between all classes/identities in the $n$-dimensional feature space. We have performed comprehensive set of experiments on a number of state-of-the-art face recognition benchmarks to confirm the efficacy of our solution. The proposed solution has consistently achieved state-of-the-art performance in all of our experiments.